# “I just assumed that it would translate”: examining MT risk awareness among healthcare staff with abbreviations as a use case

**Eleanor Taylor-Stilgoe** **Félix do Carmo** **Constantin Orăsan**

University of Surrey, Guildford, UK
{e.j.taylor-stilgoe,f.docarmo,c.orasan}@surrey.ac.uk

## Abstract

In the UK, public healthcare staff report turning to machine translation (MT) - predominantly Google Translate (GT) - to communicate with patients across language barriers. Though intended to support their duty of care, potentially uninformed reliance on MT in such contexts could have serious consequences for patient safety. Research nonetheless remains limited on staff awareness of the possible risks posed by higher-stakes MT use in general and with patient medical records in particular, most existing literature instead examining its use in interpersonal situations or with patient-oriented documentation.

Moreover, medical abbreviations are well-documented as increasing patient risk even monolingually, with outcomes from their misuse and/or misinterpretation ranging from temporary harm to the death of the patient. Abbreviations were therefore selected as a use case for identifying the potential risks posed by their translation with MT.

Contextualised French and Spanish data examples drawn from authoritative clinical corpora and translated via GT were presented during semi-structured interviews to 21 healthcare staff participants in diverse roles and specialties. The results were then subject to qualitative analysis and cross-analysis.

## 1 Introduction

In UK public healthcare, staff across diverse roles and communicative contexts strive to balance language barriers with their duty of care by turning to machine translation (MT) - predominantly Google Translate (GT) - despite official National Health Service (NHS) guidance cautioning against such use on the grounds that ‘*there is no assurance of the quality of the translations’* (2018). This decision is often driven by time pressures, financial constraints, and/or resourcing difficulties (Royal College of Midwives, 2017; Moberly, 2018; Bell et al., 2020), further underscored by a reported gulf between institutional policy and professional reality, as experienced by NHS clinicians (Vieira, 2024). Recent research by Valdez et al. (2025) further underscores this mismatch within wider practice, with MT and generative AI at a Dutch hospital being ‘*mostly used in routine care*’ but ‘*perceived as inappropriate for complex interactions, against a backdrop of absent institutional policies*’.

Furthermore, despite the centrality of medical records to patient care, research exploring the risk implications of their translation with generic MT remains limited. Recent evaluations of MT for patient-specific discharge instructions have recommended caution in practice due to risks from abbreviations, jargon, and low readability (Kong et al., 2026); the majority of the literature otherwise focuses on MT use as an interpreting aid in interpersonal encounters or for generating patient-oriented documentation, such as discharge instructions and information leaflets (Khanna et al., 2011; Patil and Davies, 2014; Chaudary and Herrera, 2016; Chen et al., 2016; Moberly, 2018; Davis et al., 2019; Taira et al., 2021).

In addition, medical abbreviations are well-documented as a major source of patient harm even at the monolingual level, with their misuse and/or misinterpretation having consequences for patient safety ranging from temporary harm to the death of the patient (Institute for Safe Medication Practices, 2001; Otero López et al., 2004; Brunetti et al., 2007; BBC News, 2008; Sheppard et al., 2008; The Joint Commission, 2026). In spite of the potential for risk amplification arising from the combination of abbreviations with a second language and generic MT, this area remains underexplored, particularly within the context of patient medical record documentation.

Against this backdrop, this study seeks to explore two critical aspects: firstly, the extent to which healthcare staff are aware of the potential risks posed to patient safety when using MT to translate patient medical record documentation; secondly, those risks identified in the use of medical abbreviations when translated with MT.

The study employs contextualised examples of abbreviations drawn from authoritative French and Spanish clinical corpora. These serve as a focused use case with which to identify both those risks arising from their translation with GT - given its prevalence in the literature - and the extent to which healthcare staff understand said risks.

## 2 Related work

MT literacy, defined by Bowker and Buitrago Ciro (2019, cited in Vieira et al., 2022) as ranging from '*a basic understanding of how MT works and the ability to judge when MT systems can be used to knowing what content should be translated with them and how their output can be improved*', remains underexplored in higher-stakes domains. Further to this, Vieira et al. (2022) surveyed 1200 UK residents, finding MT primarily used for leisure (80.1%) in lower-stakes contexts; only 5.2% of those expressed risk concerns in a medical context. Somewhat alarmingly, one healthcare professional reported having used GT for medication administration purposes despite being unable to recognise potential translation errors due to source language (SL) unfamiliarity (2022). Additionally, non-translator users of MT have previously been documented as conflating accuracy with grammatical fluency rather than contextual fitness for purpose (Martindale and Carpuat, 2018; Robertson and Díaz, 2022). In the context of patient medical record documentation, such misjudgements could put patients at risk.

Despite the prior development of domain-specific tools (Starlander et al., 2005; Albrecht et al., 2013; Bouillon et al., 2017; Noack et al., 2020), healthcare staff reportedly turn to MT – predominantly GT - amid inadequate access to language services, time pressures, and/or cost constraints. These difficulties are further highlighted by the apparent chasm between institutional ideals and the actual practice conditions experienced by clinicians (Vieira, 2024). GT therefore appears to be viewed as a 'better than nothing' solution by healthcare staff (Royal College of Midwives, 2017; Moberly, 2018; Bell et al., 2020; Vieira et al., 2020). Recent international research reinforces this pattern among healthcare professionals, with staff turning to MT for routine care but not for more complex scenarios, alongside a lack of institutional policies or training (Valdez et al., 2025).

In addition, medical abbreviations – a ubiquitous yet error-prone feature of the medical domain - contribute to the potential for patient harm through misuse and/or misinterpretation. The documented consequences of such abbreviations in medical contexts range from temporary harm to the patient's death (Institute for Safe Medication Practices, 2001; Otero López et al., 2004; Brunetti et al., 2007; Sheppard et al., 2008). Abbreviation knowledge is documented as varying across healthcare professionals (Das-Purkayastha et al., 2004; Chemali et al., 2015; Awan et al., 2016), the risk of misinterpretation being further amplified by a lack of consistent standardisation (The Joint Commission, 2026). While recent evaluations of MT for patient-specific discharge instructions have served to highlight how abbreviations aggravate risks (Kong et al., 2026), the translation of medical abbreviations in the context of patient medical record documentation remains an overlooked subdomain in an already understudied area.

## 3 Data research methodology

### 3.1 Data collection

Medical abbreviations - defined broadly per Sheppard et al. (2008) as '*any shortened form of a word* […] *any acronym* […] *any contracture* […] *and any initialism*' - were selected as the use case with which to examine risk creation or amplification when translated into English with MT. Authentic clinical corpora in French and Spanish were selected to ensure authoritative, reliable data that closely reflected the nature and linguistic features of real-life documentation.

First selected was the French CAS corpus, comprising 4,900 clinical cases drawn from legal and medical training resources, as well as scientific literature (Grabar et al., 2020). This included both real (de-identified) and fake patient cases. The second corpus selected was the Spanish CodiEsp clinical case study corpus (Miranda-Escalada et al., 2020), comprising a randomly sampled development dataset (250 cases), test set (250 cases), training set (500 cases), and an unannotated background set (2,751 cases). Both corpora were selected for their coverage of various medical specialties, patient comorbidities and expert-to-expert case histories, echoing elements encountered in authentic medical records.

### 3.2 Data preparation and processing

A custom desktop application was designed by the first author and implemented by a professional software developer in C# and using Microsoft Visual Studio. This application supported configurable sentence filtering using regular expressions, manual exclusion of irrelevant matches, and exporting selected matches to an Excel file for subsequent translation with GT and analysis in Sketch Engine (Kilgarriff et al., 2014). Noise in the source text (ST) data (e.g., spelling/grammatical errors) was intentionally preserved to reflect realistic MT input conditions. The paragraph in which each match occurred (as delimited by line breaks) was also preserved to ensure sufficient SL context for fairer and more realistic assessment by healthcare staff.

A fuller description of the custom extraction tool is available in the first author's doctoral thesis (Taylor-Stilgoe, 2026).

### 3.3 Data example selection

A random set of SL matches were analysed to explore any issues arising from their translation with GT. Two particularly salient examples were selected for use in the study as a result. The first was an ophthalmological Spanish text containing four untranslated instances of the abbreviation 'AO' ('ambos ojos' / 'OU', 'oculus uterque' or 'BE', 'both eyes'), in the context of a clinically normal eye examination; the second was a primarily cardiological French text containing a single but contextually key instance of the abbreviation 'DRS' ('douleurs rétrosternales' / 'retrosternal pain'), reflecting ongoing retrosternal pain symptoms despite the patient having undergone numerous prior investigations. The full examples are provided in Appendix A.

## 4 Interview methodology

### 4.1 Participant recruitment

Participants approached for recruitment included clinical, clinical support (i.e., healthcare assistant) and non-clinical roles. This professional diversity potentially informs the type of information each participant might seek to identify in the ST and/or MT, their degree of clinical risk awareness, and their approach to risk management. No specific criteria were assigned concerning participants' language knowledge, length of service or specialty to mirror the diversity of UK healthcare settings and potentially yield richer analyses as a result. Participants were recruited via snowball sampling.

Of the 21 participants recruited, 15 performed clinical roles, 2 clinical support roles, and 4 non-clinical roles, all but one of whom worked in NHS settings. Their full roles are listed in Appendix B.

### 4.2 Study design and implementation

The interview process consisted of three stages, the full list of questions for which is provided in Appendix C. Interviews were conducted remotely and lasted an average of 90 minutes, allowing for extensive exploration of participants' views.

During Stage 1, participants answered open questions about their professional background, attitude towards, and practices with MT in lower-stakes settings (e.g., personal usage for holidays) and higher-stakes contexts respectively. It was essential to start with this stage to safeguard against priming or social desirability bias before exposing participants to the data examples.

These examples were then presented to participants in structurally identical stages, with the Spanish 'AO' appearing in Stage 2 and the French 'DRS' in Stage 3, this sequencing reflecting an intentional escalation in possible severity (i.e., ophthalmology with a clinically normal outcome followed by primarily cardiology with ongoing symptoms). Each was presented on a blind basis to avoid priming and contextualised for a scenario typical of each participant's work environment to maximise response authenticity. Stages 2 and 3 were structured as follows:

1. Participants were first asked what information they would look for in the ST and why **(Q1)** with a view to observing their priorities (e.g., specialty, numbers/units) and whether or not these included the use case in question.

2. Participants were then shown the corresponding English MT (retaining the ST) and asked what action they would take on identifying a possible error in the output **(Q2)**. If no issues were spotted, participants could answer on a more general basis to explore their typical MT risk management practices.

3. Participants then scored their degree of general confidence in using the MT text in a real-life context and were asked to give the reason for their score **(Q3a)**, using a 6-point Likert scale: 0 = not at all confident; 1 = slightly confident; 2 = somewhat confident; 3 = fairly confident; 4 = confident; 5 = completely confident.

4. Participants then scored the potential for patient harm posed by use of the MT text and were asked to give a reason for their score (**Q4a**). This was scored using a 4-point scale adapted from the National Coordinating Council for Medication Error Reporting and Prevention (NCC MERP) Index for Categorizing Medication Errors (Brunetti et al., 2007), as follows: 0 = no capacity to cause harm; 1 = capacity to cause harm; 2 = capacity to contribute to or result in temporary harm requiring intervention; 3 = capacity to contribute to or result in death.

These scoring evaluations were then followed by a discussion stage covering the general content of the MT (e.g., the specialty, scenario, and the nature of the information being conveyed), the meaning and clinical significance of the key problematic SL abbreviation, its correct English translation, and the possible safety implications of its being left untranslated in the MT text. Participants were invited to share their thoughts on both these elements and any other phenomena, given the scope this provided for gaining real-time insight into any changes in their MT risk awareness during the discussion stage.

Following discussion, participants were asked to provide another score and the corresponding reason(s) for both general confidence **(Q3b)** and patient harm **(Q4b)**.

Lastly, participants were invited to give optional feedback on what they had learned or any other points relevant to the research aims.

### 4.3 Data analysis framework

In terms of data analysis, a higher-level, conceptually-based coding structure reflecting the questions posed during Stages 1, 2, and 3 was established at the outset, with hierarchical subcodes being added pursuant to the responses provided. These responses were then subject to qualitative content analysis in MAXQDA 2022. Additionally, data-driven insights of relevance to the research aims were hierarchically coded by theme for subsequent thematic analysis.

It should be noted that, for Stage 1, the subcodes to which lower-stakes MT attitudes were assigned comprised only 'negative', 'neutral', and 'positive'. However, in order to accurately classify attitudes towards higher-stakes use, it was necessary to create a 'pragmatic' subcode that reflected participants' 'better than nothing' use of MT at work, irrespective of their personal attitudes.

## 5 Findings and discussion

### 5.1 Stage 1: Attitude towards MT use across lower- and higher-stakes settings (n=15)

A notable cross-stakes attitude shift was observed among 93% of participants providing comparable responses. In lower-stakes settings, attitudes reflected either indifference towards or active willingness to use MT. However, these attitudes turned sharply towards the negative in higher-stakes settings. Whereas lower-stakes settings yielded 60% neutral and 40% positive attitudes towards MT use, these attitudes became 20% neutral or neutral pragmatic, 27% positive or positive pragmatic, and 53% negative or negative pragmatic, in higher-stakes settings (Figure 1).

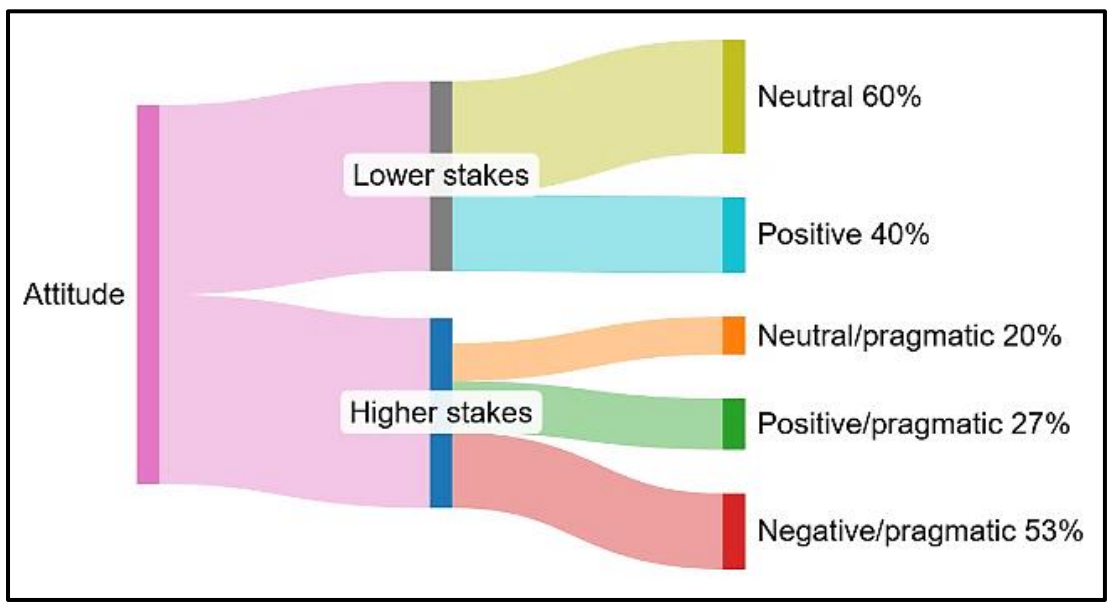

Figure 1: Attitude towards cross-stakes MT use (n=15).

The grounds for participants' reluctance or outright refusal to use MT with patient medical record documentation ranged from concerns around data confidentiality, accountability, liability, and (non-) compliance with official policy to fears of jeopardising patient safety through recognition of the technology's limitations based on prior personal and/or professional experience. Moreover, those participants specifically demonstrating a negative pragmatic attitude reported that they would turn or had turned to GT as a 'last resort', despite their personal misgivings, for want of a viable alternative in practice.

A clear demonstration of this shift came from one non-clinical participant's negative higher-stakes attitude, derived from their experience of using GT with their own medical records. On this point, they asserted that "[...] *I wouldn't feel comfortable using* [GT] *for patients because it's not 100% accurate. Some of the stuff from my own experience* [...] *was coming up to something completely different or it didn't make sense. It would be like putting* [the] *patient at risk*" (IV9, Cardiology Administrator). Additionally, one clinician who reported previously having to triage children in four minutes - this including any translation needs - adopted a negative pragmatic attitude to using MT, commenting that "*I guess 'willing' is probably... I don't know. I think it's a case of having to, there aren't any other options, if you know what I mean?*" (IV11, Children's Nurse).

Moreover, two clinical participants - one of whom demonstrated a positive attitude and the other a positive pragmatic attitude - avoided overreliance on MT in isolation by only using the output as guidelines for integration into the wider clinical picture. As such, their attitudes suggested at least a degree of awareness of the risks involved.

## 5.2 Stage 2: Impact of 'AO' on general confidence scoring (n=21)

During Q1, no participants referred to 'AO' despite 48% mentioning other abbreviations present in the ST. For Q3a, of the 24% participants citing abbreviations as an influence on their general confidence in using the text, only 5% cited 'AO'. During Q3b, however, following discussion of the issues present, 'AO' was cited by 19% of participants, representing an overall influence increase of 14% (Figure 2).

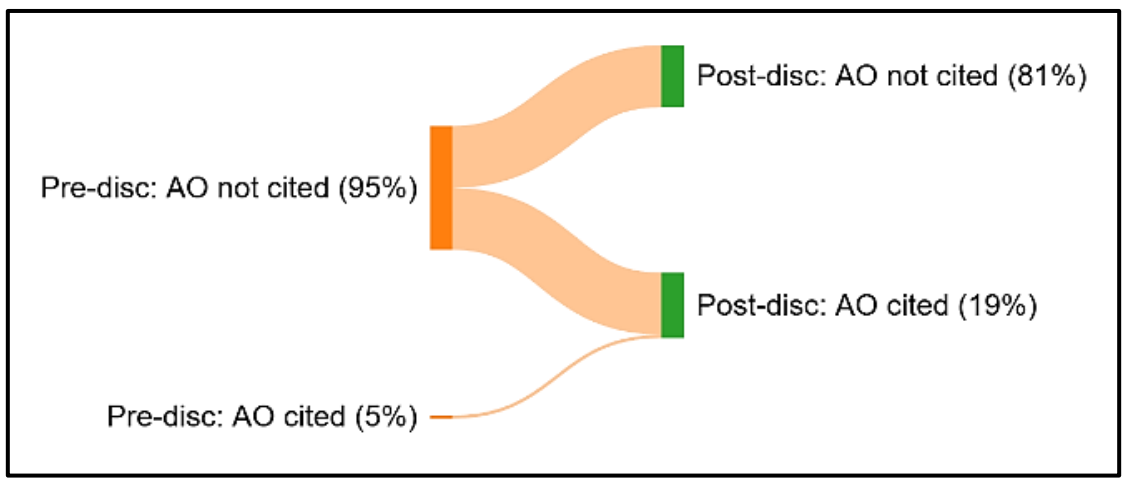

Figure 2: Impact of 'AO' on general confidence scoring, before and after the discussion stage (n=21).

A particularly illustrative case was one clinician reducing their score from 4 ('confident') to 3 ('fairly confident') during Q3b, having learned the meaning of 'AO'. On identifying all four instances in the MT, they reflected on the loss of clinically relevant information through non-translation. This would be especially problematic if 'AO' were also unidentifiable by those in the specialty: "[...] *even though it might not be something I'm sure about, if it goes to the specialism and they don't know what that is as well, that's quite a bit of key information missing*" (IV10, Nurse). Another clinician working in Accident and Emergency (A&E) reduced their score from 4 to 3 after reflecting on their assumptions about the significance of 'AO', especially when scanning the MT under time constraints: "[...] *I kind of whizzed through* [the abbreviations] *and thought, 'I'm not going to know in a time-pressured environment what that is, but if I can work out the important stuff from this, I can* [...] *just ignore that', whereas now I'm like, 'Ok, okay,* ['AO'] *is really important, it came up quite a few times* [...]*"* (IV15, Junior Doctor).

### Stage 2: Impact of 'AO' on patient harm scoring (n=18)

During Q4a, no participants referred to abbreviations as influencing their risk perception. During Q4b, however, 17% indicated that learning the meaning of 'AO' had negatively impacted their perception of the degree of risk posed. This represents an overall influence increase of 17% (Figure 3).

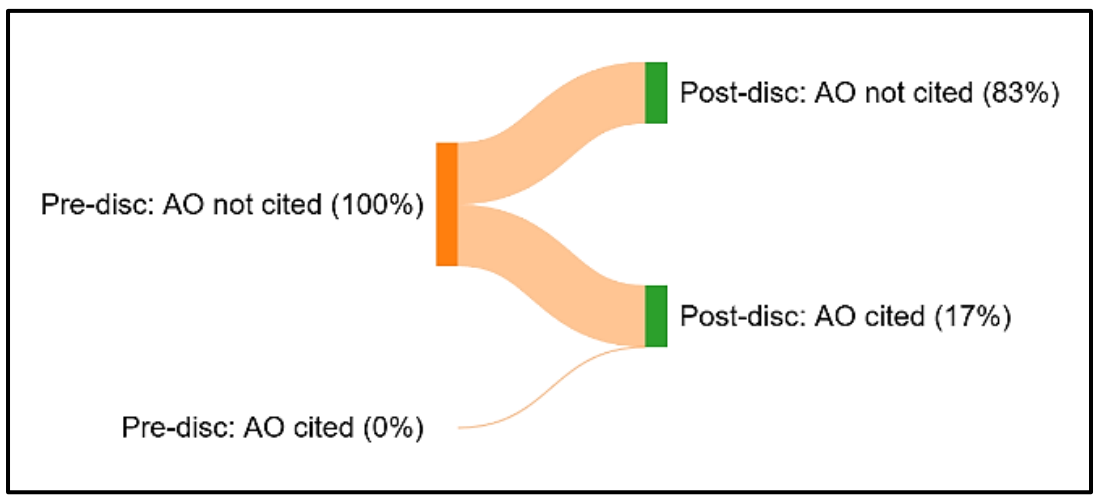

Figure 3: Impact of 'AO' on patient harm scoring, before and after the discussion stage (n=18).

One clinician's post-discussion score change drew attention to the risks inherent in expecting unknown abbreviations to be recognisable to the specialty. On raising their score from 0 ('no capacity to cause harm') to 2 ('capacity to contribute to or result in temporary harm requiring intervention'), they reflected on their prior assumption that those in the specialty would be able to identify 'AO': "[...] *there's more risk now I know this than I thought initially* [...] *there was a lot of assumption that somebody else would know but, actually, they might not* [...]" (IV21, Advanced Clinical Practitioner).

### 5.3 Stage 3: Impact of 'DRS' on general confidence scoring (n=21)

During Q1, only 19% participants made reference to 'DRS', despite its being the sole abbreviation present in the ST. For Q3a, only 5% cited 'DRS' as an influence, with some participants interpreting 'DRS' as 'doctors', '[health] services', or 'Emergency Department' rather than symptoms. During Q3b, however, 52% of participants cited the potential risks posed by the untranslated 'DRS' as a contributor to their score change, representing an overall influence increase of 47% (Figure 4).

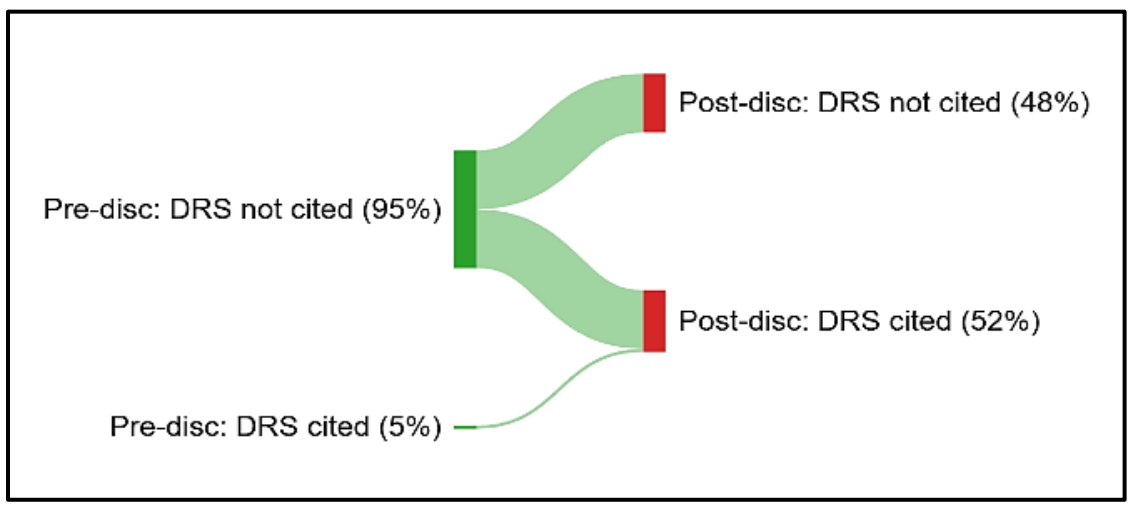

Figure 4: Impact of 'DRS' on general confidence scoring before and after the discussion stage (n=21).

This shift was vividly captured by one non-clinical participant whose confidence plummeted following discussion, reducing their score from 4 ('confident') to 2 ('somewhat confident'). The non-translation of 'DRS', particularly in the context of unresolved chest pain symptoms, had reportedly not only shaken their risk perception concerning the use case but also their confidence in the MT's reliability as a whole: "[...] *that took my confidence completely away* [...] *it makes sense, what is above, but does it?* [...] *I don't trust it anymore*" (IV1, Referral Hub Supervisor).

### 5.4 Stage 3: Impact of 'DRS' on patient harm scoring (n=18)

During Q4a, 22% of participants cited 'DRS' as an influence on their score. By contrast, during Q4b, 33% cited 'DRS' as an influence on their risk assessment, representing an overall influence increase of 11% (Figure 5).

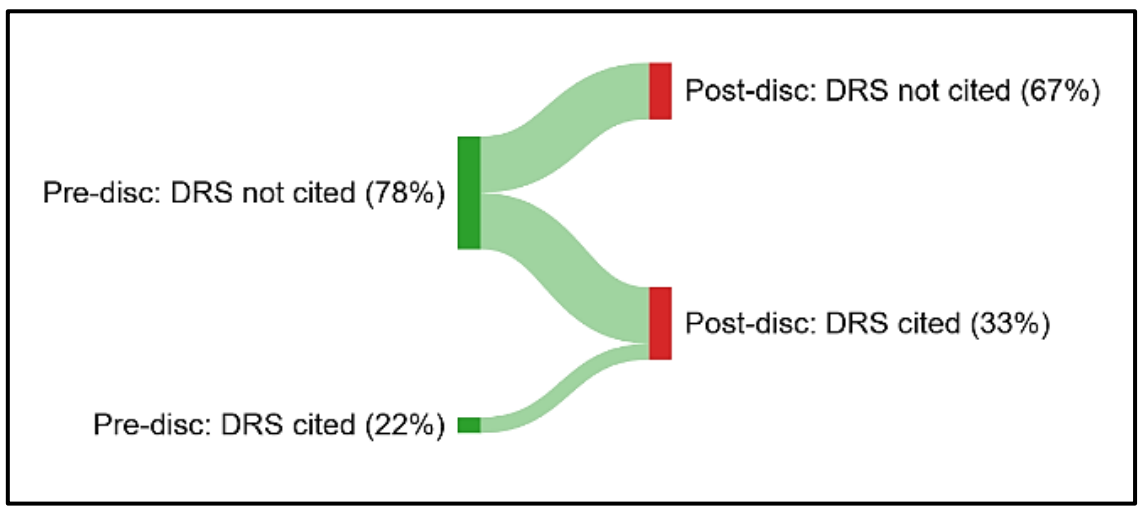

Figure 5: Impact of 'DRS' on patient harm scoring, before and after the discussion stage (n=18).

A thematically illuminating case was one clinician's increase in score from 1 ('capacity to cause harm') to 2 ('capacity to contribute to or result in temporary harm requiring intervention') due to treatment delay concerns, the participant themselves working in time-critical care. Whilst confident about obtaining symptom information via other means, they considered the time lost in doing so to indirectly contribute to risk creation: "[...] *you would have to be doing the gestures and things, so it would slow the examination down*" (IV19, Stroke Nurse Specialist).

The overall impact of 'AO' and 'DRS' on participants' general confidence (GC) and perception of the potential for harm arising from its use (PH) is summarised in Table 1 below.

| Use case | Score type | Before | After | Increase |
|---|---|---|---|---|
| AO | GC | 5% | 19% | +14% |
| AO | PH | 0% | 17% | +17% |
| DRS | GC | 5% | 52% | +47% |
| DRS | PH | 22% | 33% | +11% |

Table 1: Impact of 'AO' and 'DRS' on participant scores, before and after the discussion stage.

### 5.5 Summary of key findings: healthcare staff risk awareness in using MT with patient medical record documentation

Further to the attitude shift described in Section 5.1 above, no participants would rely on the MT's fitness for purpose in isolation, instead deploying various risk management strategies and compensation techniques to mitigate the potential risks posed by MT ambiguities and errors. These strategies included integration of the MT into the wider clinical assessment, double-checking the text with an available bilingual speaker, referring it to a managerial senior or clinical specialist, cross-referencing MT ambiguities online or, where feasible, referring to the patient for contextual confirmation.

However, some responses suggested the possible conflation of linguistic quality (e.g., fluency/readability, semantic precision) with translation quality, with the former effectively being treated as a proxy for evaluating MT fitness for purpose. A small number of responses also suggested the conflation of bilingualism with domain-specific translation competence, including among those with prior linguistic knowledge. While the decision to refer the ST and/or MT to a bilingual speaker may stem from practical constraints, such practices nonetheless carry inherent risks if rooted in incomplete understanding of the same.

Of particular concern was the frequency with which participants failed to notice, overlooked or misinterpreted the use case abbreviations, these oversights occurring across roles, length of service, language knowledge, and prior exposure to MT. This risk was further underscored by various participants' gravitation towards Romance cognate-friendlier elements (e.g., 'TAC'/'CAT [scan]') and/or English-like elements in the ST (e.g., 'RBC'/'red blood count') at the expense of less readily transparent elements like 'AO' and 'DRS'. Where overlooked, this was often due to the assumption that they belonged to and would thus be recognisable by another specialty; in other cases, they were misinterpreted as other information (e.g., 'AO' as medication, 'DRS' as 'doctors'), leading to a loss of clinically significant information. As noted by one clinician, "*It's interesting how it's this one thing, just one abbreviation of this entire thing, that leaves you in the dark with how to treat the patient, basically*" (IV8, Stroke Nurse Specialist).

### 5.6 Summary of key findings: potential risks identified through the use of abbreviations with MT in healthcare

Among the most notable findings were the potential risk implications arising from use of MT, should the use case abbreviations remain unresolved or be subject to misinterpretation. Participants determined risks to range from the prescription of inappropriate and possibly harmful medications to treatment delays, the delivery of inappropriate treatment, and failure to provide necessary treatment. The corresponding clinical outcomes were determined to range from temporary harm to the death of the patient.

Additionally, a number of clinical participants reported regularly encountering insufficiently contextualised and/or non-standard abbreviations in the workplace, these habits reportedly creating confusion at the monolingual level alone. Failure to provide the full preceding term at least once in the ST was therefore identified as a key contributor to risk creation, in depriving the MT of contextual clarity at the earliest point in written communications.

As such, medical abbreviations proved an effective use case for exploring the risks posed by their (non-) translation with MT, highlighting both the ripple effect and possible clinical outcomes of inadequate source contextualisation.

### 5.7 Feedback review: impact of the study on participants' views (n=18)

Of those participants offering comments relevant to the study, 22% expressed concern about the impact of language barriers on patient care and, in some cases, their hopes for the study to contribute to addressing this issue. As one clinician working in A&E noted: "[...] *it is an area that needs to be improved for us* [...] *given that it's so easy to go from one country to the other, that, actually, we need to have the ability to treat all* [...]" (IV13, A&E Registrar).

Furthermore, 39% had been inspired to reflect on their professional practices, with one clinician potentially adjusting their future abbreviation usage despite time constraints: "*We use abbreviations all the time at work* [...] *if I was to document, should I be using abbreviations, or should I be using the full text? And I think I should be using the full text. But time is a factor* [...] *Abbreviations in themselves are a different language*" (IV2, Advanced Clinical Practitioner).

Finally, 50% of commenters expressed a shift in MT risk awareness as a result of the interview study. For instance, one clinician who asserted having "[...] *just assumed that it would translate*" reflected on their confidence in information chains based primarily on collegial trust: "*This actually made me think, 'Is that a good translation?'* [...] *I may make decisions on what that Portuguese nurse has told me, and it could be wrong. Maybe she didn't read it right*" (IV8, Stroke Nurse Specialist). Another clinician reported having been inspired to reflect on their trust in MT outputs potentially containing subtler errors, stating that "[...] *I don't think I would trust* [GT] *as much as I would have normally. And, yeah, just always get stuff checked, because you just never quite know*" (IV23, Adult Nurse).

## 6 Conclusions and recommendations

While participants showed largely sound awareness of MT's general limitations when applied to a higher-stakes context, gaps persisted in their recognition of subtler errors. For example, untranslated medical abbreviations were presumed to belong to another specialty, represent a different target concept entirely, or were glossed over in favour of more transparent Romance cognates and/or English-like elements. Moreover, the study revealed that abbreviations devoid of the full term at least once at source pose a potentially amplified risk when run through MT, as the lack of context directly impacts their (un)successful translation. The severity of the possible consequences for patient safety, as identified by the study participants, underscores how even seemingly minor untranslated elements, such as a single abbreviation, can cascade into substantial clinical consequences in higher-stakes settings.

Based on these findings, healthcare professionals are advised to exercise caution when using MT to translate patient medical record documentation, especially given the prevalence of abbreviations in this context. STs should also accompany the MT during double-checks to identify abbreviation-related (or other) translation issues as promptly and effectively as possible.

Furthermore, given the volume and diversity of abbreviations used in healthcare in general and medical records in particular, omission of the full preceding term at least once at source has the very real potential to contribute to risk creation, especially when coupled with insufficient MT-related risk awareness. It is therefore essential to ensure the inclusion of the full term at least once to reduce the risk of downstream translation errors. By way of supportive measures, a context-sensitive abbreviation checker could be developed for integration into EHR systems. This approach would enable better safeguarding of record accuracy without unduly increasing manual reporting burdens on already pressured staff.

In addition, targeted training designed to enhance MT literacy and risk awareness among healthcare staff would help to address possible knowledge gaps around translation quality, translation competence, and the MT-related risks identified in this study. Such measures would both build staff confidence in better understanding the strengths, weaknesses, and risks involved in using MT at work and mitigate the risk of patient harm arising from its inadequately informed use.

## 7 Positionality statement

The first author brings approximately four years of professional healthcare experience and eight years in freelance translation to this work. These roles informed the identification of medical abbreviations as a higher-risk MT use case, aided understanding of the pressures on staff, and shaped the design of contextualised, role-specific interview scenarios to elicit more authentic participant insights. While this insider perspective enhances the depth and relevance of the findings, strict controls were put in place to maintain researcher neutrality and mitigate potential bias.

## 8 Limitations of the study

A notable limitation was the restriction of the SLs to French and Spanish. While including scripts relevant to major UK migrant communities would have further enhanced analysis, this was unfeasible due to limited domain-specific data and the resources required to validate MT outputs.

A further limitation was the small sample size and use of snowball sampling, restricting generalisability and risking sample homogeneity. While these constraints reflect the challenges of recruiting healthcare professionals facing heavy workloads, accessing this population even in a modest capacity yielded rich and contextually grounded insights. This demonstrates the value of qualitative research despite practical barriers and suggests possible future avenues for studies using larger and/or more linguistically diverse samples.

## A Appendix

### 'AO' and 'DRS' data examples

The full use case data examples are provided below (emphasis and/or [sic] added for clarity):

### Example 1: 'AO'

ST: *En el examen inicial presentaba una AV con su corrección (AV CC) de 0,2 en* ***AO*** *que mejoraba a 0,4 con estenopeico. La motilidad pupilar era normal, la presión intraocular (PIO) de 18mm Hg en* ***AO*** *y no presentaba alteraciones en la biomicroscopia anterior (BMA). La refracción bajo cicloplejía (RBC) era de -4,50 en* ***AO****. En el fondo de ojo (FO) se apreciaban estrías retinianas en la mácula de* ***AO****. La tomografía axial computarizada (TAC) no reveló anomalías en la región orbitaria ni asimetrías ni alteraciones en la posición del diafragma iridocristaliniano.*

GT: *In the initial examination, she presented a VA corrected for her (VA CC) of 0.2 in* ***AO*** *that improved to 0.4 with pinhole. Pupillary motility was normal, intraocular pressure (IOP) was 18mmHg in* ***AO****, and she had no alterations in anterior biomicroscopy (AMB). Refraction under cycloplegia (RBC) was -4.50 in* ***AO****. Retinal striae were seen in the eye fundus (FO) in the* ***AO*** *macula. Computerized axial tomography (CAT) revealed no abnormalities in the orbital region or asymmetries or alterations in the position of the iridocrystalline diaphragm.*

This Spanish to English ophthalmological example contains four instances in which the abbreviation 'AO' ('ambos ojos' in Spanish, for which the correction translation would be the Latin 'OU', 'oculus uterque' or English 'BE', 'both eyes') remains wholly untranslated in the MT, seemingly due to the lack of a full term preceding the SL abbreviation. The result is the omission of central clinical information as to the relevant examination having been conducted on both eyes. While other abbreviations present in the MT, such as 'RBC' ('refracción bajo cicloplejía'/'cycloplegic refraction', 'CRx') and 'FO' ('fondo de ojo'/'[eye] fundus') remain either mistranslated or untranslated, these are at least preceded by the full if only broadly rendered terms, whereas this surrounding context remains absent for 'AO'.

### Example 2: 'DRS'

ST: *Au cours des dix dernières années, elle a subi quatre coronarographies, neuf échographies cardiaques et douze épreuves d'effort au tapis roulant. Depuis la dernière dilatation, quatre ans auparavant, aucune lésion significative ou donnée probante de resténose ne sont notées aux coronarographies. Les pressions pulmonaires sont normales. Aucune maladie valvulaire n'apparaît aux échographies. Les paramètres électrophysiologiques en relation avec les suivis liés à son stimulateur cardiaque défibrillateur sont adéquats* [sic]*; aucune anomalie significative n'apparaît et aucune thérapie ne lui est donc appliquée. Bien que cette dame ne soit pas suivie dans cet établissement, elle nous assure que son lupus est stable. D'ailleurs, la protéine C réactive est normale, la tension artérielle est bien contrôlée, et les douleurs angineuses n'ont pas un caractère péricardique. Enfin, elle n'est connue pour aucun antécédent psychiatrique. Pourtant, la patiente continue de présenter des* ***DRS****.*

GT: *Over the past ten years, she has had four coronary angiograms, nine cardiac ultrasounds and twelve treadmill stress tests. Since the last dilation, four years earlier, no significant lesions or evidence of restenosis have been noted on coronary angiograms. Pulmonary pressures are normal. No valvular disease appears on ultrasound. The electrophysiological parameters in relation to the follow-ups related to his defibrillator pacemaker are adequate; no significant abnormality appears and no therapy is therefore applied to him. Although this lady is not followed in this establishment, she assures us that her lupus is stable. Moreover, C-reactive protein is normal, blood pressure is well controlled, and angina pain is not pericardial in nature. Finally, she is not known to have any psychiatric history. However, the patient continues to present with* ***DRS****.*

This largely cardiological French to English example, which shifts to psychiatry in the penultimate sentence, reflects a complete failure by the MT to translate 'DRS' ('douleurs rétrosternales'), for which the correct translation would be 'retrosternal pain'. While this non-translation also appears to have arisen from the lack of a full term preceding the SL abbreviation, no direct equivalent exists in English for 'retrosternal pain'. Despite being the only abbreviation in the paragraph, 'DRS' is of central importance to identifying the patient's ongoing symptoms.

## B Appendix

**Participants' identifiers and respective professional roles are listed below, including those former roles relevant to their interview responses. IV4 and IV14 were unable to attend.**

| ID | Professional role | Nature of role |
|---|---|---|
| IV1 | Referral Hub Supervisor<br>Former A&E coordinator | Non-clinical |
| IV2 | Advanced Clinical Practitioner<br>Former emergency care nursing roles | Clinical |
| IV3 | Healthcare Assistant, Sexual Health | Clinical support |
| IV5 | Podiatry Administrator | Non-clinical |
| IV6 | Healthcare Assistant, Mental Health (private care) | Clinical support |
| IV7 | Cardiology Booking Coordinator | Non-clinical |
| IV8 | Stroke Nurse Specialist | Clinical |
| IV9 | Cardiology Administrator | Non-clinical |
| IV10 | Nurse | Clinical |
| IV11 | Children's Nurse | Clinical |
| IV12 | Trainee Advanced Clinical Practitioner<br>Nurse (including military nursing) | Clinical |
| IV13 | Registrar, A&E | Clinical |
| IV15 | Junior Doctor, A&E | Clinical |
| IV16 | Clinical Psychologist (community neurological services, Weight Management, and Sexual Health Services teams) | Clinical |
| IV17 | Dietitian | Clinical |
| IV18 | Obesity Dietitian | Clinical |
| IV19 | Stroke Nurse Specialist | Clinical |
| IV20 | Highly Specialist Neurological Physiotherapist (community services) | Clinical |
| IV21 | Advanced Clinical Practitioner (General Practice) | Clinical |
| IV22 | Nurse | Clinical |
| IV23 | Adult Nurse<br>Former Healthcare Assistant | Clinical |

## C Appendix

### Stage 1 interview questions

**Lower-stakes settings**

Do you ever use MT in your personal life? If so, which one(s), and what do you think of it?

**Higher stakes settings**

How long have you been in your current professional role? Tell me a little bit about what you do.

In the course of your current professional role, have you ever experienced a situation in which you or a colleague needed to understand patient medical record documentation written in a foreign language, but couldn't? If so, how did you manage it, and what was the outcome?

How frequently do you come into contact with patient medical record documentation written in a foreign language, in what type of situations, and for what purposes?

Have you, do you, or would you be willing to use a machine translation engine, such as Google Translate, to help you manage patient medical record documentation written in a foreign language? If not, why not?

### Stage 2 and 3 interview questions

**Presentation of the ST only**

**Q1.** This document is just an example of patient medical record documentation written in [Spanish/French]. Imagine if you were in [contextualised situation] and were presented with this document. What kind of information in the documentation, if any, would you be looking to identify, and for what purposes?

**Presentation of the MT with ST available for reference**

**Q2.** If you identified, or thought you may have identified, an error in the machine translation output, what would you do?

**Confidence scoring requested prior to and following the discussion stage**

**Q3a/b.** How confident would you feel about using this machine translation output in a real-life professional situation? Please give a score for your degree of confidence and explain your answer.

**Potential for patient harm scoring requested prior to and following the discussion stage**

**Q4a/b.** Please provide a score for the degree of potential risk to patient safety you would consider this machine translation output to pose if used in a real-life professional situation. Please explain your answer.